\documentclass[10pt,letterpaper,twocolumn]{article}
\usepackage[letterpaper,top=0.78in,bottom=0.79in,left=0.76in,right=0.76in]{geometry}
\usepackage[T1]{fontenc}
\usepackage{times}
\usepackage{microtype}
\usepackage{amsmath,amssymb,amsfonts}
\usepackage{graphicx,booktabs,array,tabularx,xspace}
\usepackage[font=small,labelfont=bf]{caption}
\usepackage{algorithm,algorithmic}
\usepackage{placeins}
\usepackage{xcolor}
\usepackage[numbers,sort&compress]{natbib}
\usepackage[hyphens]{url}
\usepackage[colorlinks=true,linkcolor=black,citecolor=blue,urlcolor=blue]{hyperref}
\hypersetup{
    pdftitle={OccWAM: Occupancy World-Action Models with Joint Scene-to-Ego Flow Matching},
    pdfauthor={Qihang Huang, Wenhui Huang, Collister Chua, Shiyu Fang, Zhan Chen, Songyan Zhang, Zhidong Wang, Daiying Tian, Chen Lv}
}
\newcommand{\method}{OccWAM\xspace}
\newcommand{\Normal}{\mathcal{N}}
\newcommand{\R}{\mathbb{R}}
\newcommand{\E}{\mathbb{E}}
\newcommand{\sg}{\operatorname{sg}}
\newcommand{\cmark}{\ensuremath{\checkmark}}
\newcommand{\xmark}{\ensuremath{\times}}
\newcolumntype{Y}{>{\raggedright\arraybackslash}X}

\title{\textbf{OccWAM: Occupancy World-Action Models with Joint Scene-to-Ego Flow Matching}}
\author{
Qihang Huang\textsuperscript{1$\ast$}\quad
Wenhui Huang\textsuperscript{1$\ast\dagger$}\quad
Collister Chua\textsuperscript{1}\quad
Shiyu Fang\textsuperscript{2}\quad
Zhan Chen\textsuperscript{1}\\[3pt]
Songyan Zhang\textsuperscript{1}\quad
Zhidong Wang\textsuperscript{1}\quad
Daiying Tian\textsuperscript{3}\quad
Chen Lv\textsuperscript{1$\ddagger$}\\[7pt]
\small\textsuperscript{1}\,Nanyang Technological University, Singapore\\[1pt]
\small\textsuperscript{2}\,Tongji University, Shanghai, China\\[1pt]
\small\textsuperscript{3}\,Agency for Science, Technology and Research (A*STAR), Singapore
}

\date{}

\begin{document}
\maketitle

\begingroup
\renewcommand{\thefootnote}{\fnsymbol{footnote}}
\footnotetext[1]{Equal contribution.}
\footnotetext[2]{Project Lead.}
\footnotetext[3]{Corresponding author.}
\endgroup

\begin{abstract}
Occupancy world models provide a structured, metric representation of future driving scenes, making them a promising basis for end-to-end autonomous driving. However, existing occupancy-aware planners typically couple occupancy prediction and action generation through either hierarchical coupling or parallel multi-task learning. Hierarchical methods use occupancy either as a fully predicted future preceding action generation or as an action-conditioned rollout for trajectory evaluation, whereas parallel multi-task methods predict occupancy alongside planning through a separate branch or auxiliary objective. These interfaces retain either an explicit generation order or a role asymmetry, limiting reciprocal interaction between world prediction and action generation during trajectory formation. We propose \method, an occupancy-native world--action model that jointly generates dense future occupancy and continuous ego trajectories via flow matching conditioned on scene representations. Building on this joint formulation, we introduce scene-to-ego world--action flow matching, a two-stage formulation. The first stage predicts scene-centric occupancy together with an absolute ego trajectory, providing a stable representation of future scene evolution. The same predicted world is then re-indexed into the future ego frames induced by the initial trajectory, where ego motion is represented through adjacent pose increments and a second joint flow refines ego-centric occupancy and motion for action-aligned planning. Unlike video-based world--action models, \method preserves explicit semantic geometry for direct planning reasoning without pixel synthesis. Extensive experiments on Bench2Drive and nuScenes demonstrate state-of-the-art performance in closed-loop driving, open-loop planning, and future occupancy prediction.
\end{abstract}

\begin{figure}[!t]
    \centering
    \includegraphics[width=\columnwidth]{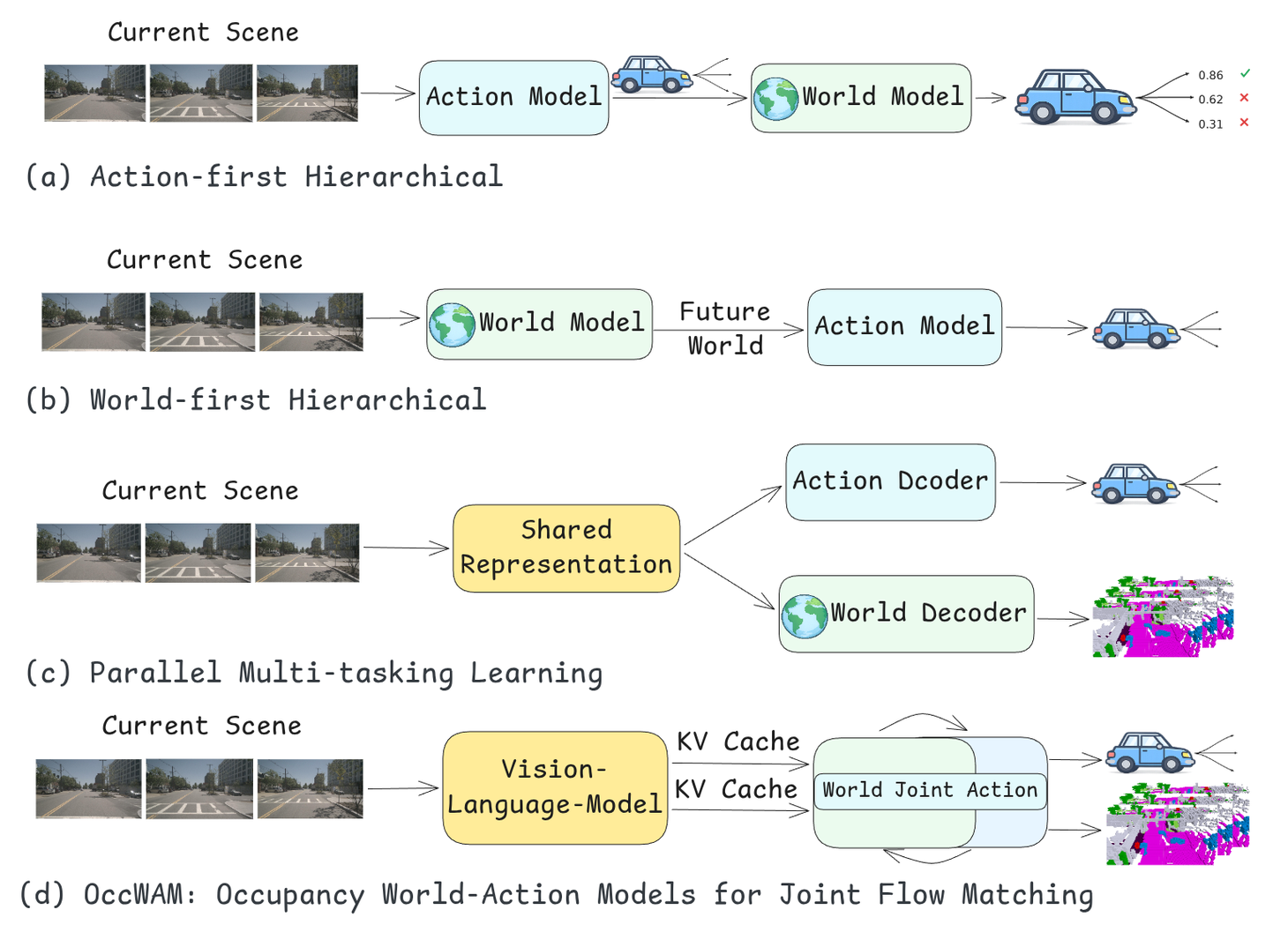}
    \caption{
    Comparison of occupancy--action interfaces.
    Hierarchical methods impose an action-first or world-first order,
    while parallel methods retain separate prediction pathways.
    \method instead jointly generates occupancy and action through
    coupled conditional flows. \emph{Multi M. Traj}: multimodal trajectory candidates.
    }
    \label{fig:teaser}
    \vspace{-5pt}
\end{figure}

\section{Introduction}

\label{sec:introduction}

End-to-end autonomous driving has progressed from direct waypoint regression toward planning-oriented structured systems, multimodal generative planners, and vision--language--action models \citep{hu2023uniad,jiang2023vad,liao2025diffusiondrive,huang2026automot,li2026unidrivevla}. These advances have substantially improved trajectory generation, task integration, and semantic reasoning. Yet a fundamental question remains unresolved: how should an explicitly modeled future participate in action generation?

A growing line of world-model-based policies and world--action models (WAMs) incorporates predictive futures into policy computation. Existing systems range from latent feature alignment and chained world-to-policy architectures to joint video--action generation \citep{zheng2025flare,ye2026dreamzero,xia2026drivelaw,shi2026drivewam}. These systems show that predicted futures benefit action generation but leave open a representation question: what form should the predicted future take to provide structured and directly usable information for planning?

Existing occupancy-based driving systems instantiate several recurring
world--action interfaces. In the action-to-world formulation, candidate actions
or intentions are generated, then their future worlds rolled out and evaluated
\citep{yang2025driveoccworld,zheng2025world4drive}. In the world-to-action
formulation, a predictive occupancy representation is constructed before being
passed to a downstream planner \citep{huang2024nemo,cheng2026owmdrive}. Others
reduce the sequential dependency via parallel multi-task decoding or ordered
discrete world--action token sequences
\citep{dang2026sparseworld,zheng2024occworld,wei2024occllama}. These
formulations retain an explicit generation order, separate output pathways, or
discrete prediction sequences. They do not instantiate the interface studied
here: dense metric occupancy and continuous ego motion jointly generated
through a coupled conditional flow.

Joint occupancy--action generation also introduces a coordinate-representation
challenge. Stable world forecasting benefits from a scene-centric frame, where
environmental evolution is not entangled with coordinate changes caused by ego
motion. Planning, in contrast, benefits from an ego-centric representation that
expresses occupied regions relative to the vehicle's intended local motion.
Scene-centric prediction has been shown to separate world evolution from
ego-induced perspective changes \citep{shi2025come}. World formation and
action-aligned reasoning therefore favor complementary rather than identical
occupancy representations.

We introduce \method, an occupancy-native world--action model that jointly
generates latent future occupancy and continuous ego motion through conditional
flow matching. World and Action Experts exchange their evolving states at every
transformer layer, allowing predicted occupancy to influence action formation
while the current action hypothesis informs future-world prediction. Occupancy
therefore participates directly in the generative planning process rather than
serving only as a completed planning condition, auxiliary objective, or
post-hoc trajectory cost. Unlike video-based WAMs, the generated world retains
explicit semantic geometry, supporting direct occupancy supervision and metric
planning queries without pixel-space synthesis.

Building on this joint formulation, we propose scene-to-ego world–action flow matching. Stage I jointly generates scene-centric occupancy and an absolute ego
trajectory in a fixed current-ego-anchored frame. The predicted world is then
re-indexed into the future ego charts induced by the Stage-I trajectory, while
the trajectory is converted into adjacent motion increments. This transformed
pair forms the structured source of Stage II, which applies a second joint
conditional flow in an action-aligned ego-centric representation. Stage I thus
models how the world and ego evolve within a stable scene frame, whereas
Stage II models how the same predicted world relates to local ego motion.

Both stages share a three-expert Mixture-of-Transformers comprising
Understanding, World, and Action Experts. The Understanding Expert provides a
frozen semantic context stream, while the World and Action Experts interact
bidirectionally throughout both flows. Sharing the backbone across both stages
keeps occupancy and action within one unified model.

We evaluate \method on complementary closed- and open-loop benchmarks.
Bench2Drive evaluates closed-loop driving under repeated interaction, while
nuScenes separately measures open-loop planning and future volumetric
occupancy, testing both the generated motion and the fidelity of the predicted
metric world.

Our contributions are threefold: (i) We introduce \method, an occupancy-native world--action model that jointly generates dense metric future occupancy and continuous ego motion through a coupled conditional flow, enabling direct bidirectional interaction between world prediction and action generation; (ii) We propose scene-to-ego world–action flow matching. A scene-centric flow first generates future occupancy together with an absolute ego trajectory; the predicted world is then re-indexed into trajectory-induced ego coordinates and jointly transported with ego pose increments by a second conditional flow; (iii) We instantiate a concrete WAM through layer-wise bidirectional interaction with a shared three-expert Mixture-of-Transformers, following the proposed scene-to-ego world–action flow matching. \method achieves state-of-the-art closed-loop driving performance on Bench2Drive and state-of-the-art open-loop planning and future-occupancy prediction on nuScenes.

\section{Related Work}
\label{sec:related_work}

\paragraph{World models and world--action models.}
World models predict future observations or scene dynamics for simulation, representation learning, and decision-making: recent work moves toward joint world--action generation in robotics \citep{zheng2025flare,zhen20243dvla,ye2026dreamzero}.
In driving, predicted futures enter action generation:
future volumetric features are forecast to inform planning
\citep{huang2024nemo}, intention-conditioned latent futures are scored
to rank trajectory candidates \citep{zheng2025world4drive},
video-generator latents condition a diffusion planner
\citep{xia2026drivelaw}, and perception and planning variables are
denoised in a shared diffusion space \citep{zhao2026united}. Closest to our formulation, DriveWAM trains video and action streams under a joint flow-matching objective over an autoregressive temporal sequence \citep{shi2026drivewam}, while generative planners model multimodal trajectory distributions without an explicit future world \citep{liao2025diffusiondrive,xu2026wamflow}. \method retains the joint objective but replaces pixel-space futures with dense metric occupancy, so the generated world is independently measurable and geometrically aligned with the
action it conditions.

\paragraph{Occupancy prediction and planning.}
Occupancy represents the future scene on a metric grid, so predictions are
evaluated directly in the space where planning constraints live. OccWorld
tokenizes semantic occupancy and autoregressively generates future scene and
ego tokens \citep{zheng2024occworld}, OccLLaMA extends this to a unified
occupancy--language--action vocabulary \citep{wei2024occllama}, and later work improves fidelity and efficiency through dynamic--static factorization \citep{zhang2024dfitoccworld}, rendering
supervision \citep{yan2025renderworld}, and next-scale temporal prediction
\citep{jin2025occtens}. Connecting the predicted world to action, a few
interfaces recur (Figure~\ref{fig:teaser}): rolling out action-conditioned
futures and selecting by occupancy cost (a) \citep{yang2025driveoccworld},
conditioning a diffusion planner on a forecast world (b)
\citep{cheng2026owmdrive}, decoding occupancy and trajectory in parallel from
shared sparse queries (c) \citep{dang2026sparseworld}, or serializing scene and
ego variables into one discrete autoregressive sequence
\citep{zheng2024occworld,wei2024occllama}. Each imposes an ordering, separate
decoders, or a discretization. Most relevant to us, COME separates
environmental evolution from ego-induced coordinate change via ego-invariant
scene-centric control features \citep{shi2025come}. \method resolves this
separation generatively: dense occupancy and continuous ego motion emerge from
a single coupled velocity field (d), with the predicted world re-indexed into
the frames its own trajectory induces, then refined jointly with the ego motion
defining them.

\paragraph{Mixture-of-Transformers.}
Mixture-of-Transformers assigns modality-specific parameters while preserving
cross-modal exchange through joint attention \citep{liang2024mot}, and couples understanding and action experts in embodied models
\citep{yu2026twinbrainvla}. Closest to our backbone, AutoMoT connects
Understanding and Action Experts through layer-wise joint attention for driving
\citep{huang2026automot}, and UniDriveVLA adds a third expert for perception
\citep{li2026unidrivevla}. \method instead makes the third expert a World Expert, which generates the future occupancy rather than parsing the present scene. World and Action Experts jointly parameterize the conditional flows of both stages on a shared three-expert
backbone.

\begin{figure*}[t]
    \centering
    \includegraphics[width=\linewidth]{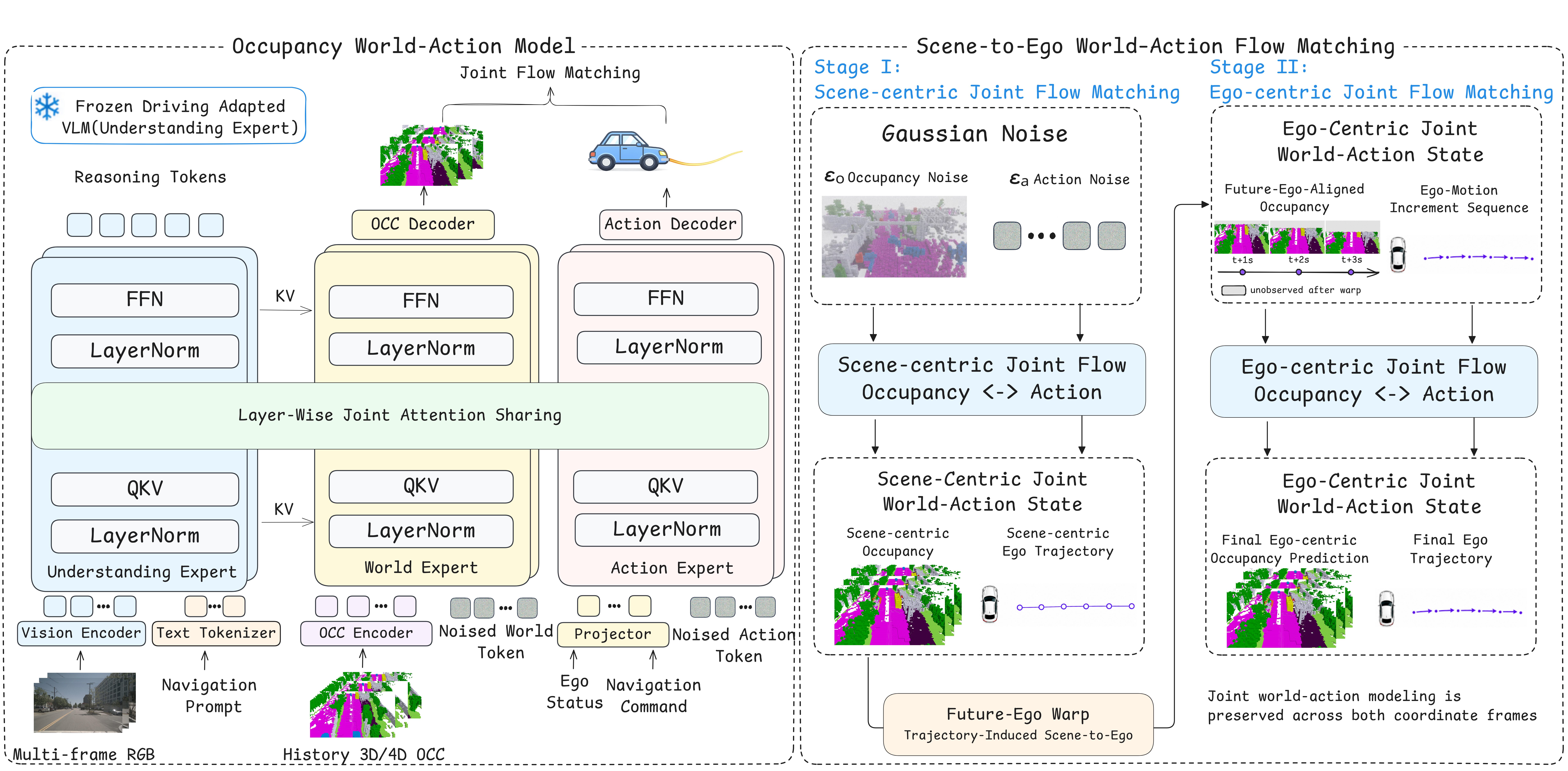}
    \caption{
    Overall architecture of \method.
    A frozen Understanding Expert provides semantic driving context, while the
    World and Action Experts interact through layer-wise joint attention to
    parameterize a coupled occupancy--action vector field.
    Stage I jointly generates scene-centric occupancy and an absolute ego
    trajectory from Gaussian sources.
    The predicted trajectory then defines future ego frames for warping the
    occupancy and is converted into ego-motion increments; together, they form the
    source of Stage II.
    Using the same three-expert backbone, Stage II applies a second joint flow in
    ego-centric coordinates to produce the final occupancy and trajectory.
    }
    \label{fig:model_architecture}
\end{figure*}

\section{Method}
\label{sec:method}

\subsection{Overview}
\label{sec:method_overview}
Given the current driving context $c$, including sensor observations, ego-motion history, navigation inputs, and benchmark-specific scene context, \method generates a paired future world and ego trajectory. Let $T$ denote the benchmark-specific number of future horizons. The Stage-I supervision is represented in a fixed scene frame as $(O^{S,\star},a^{S,\star})$, where $O^{S,\star}=\{[O^{S,\star}]_k\}_{k=1}^{T}$ is a dense semantic occupancy sequence and $a^{S,\star}\in\R^{T\times 3}$ is a cumulative planar ego-pose trajectory. Bench2Drive represents the future world as temporal BEV occupancy, whereas nuScenes uses future volumetric occupancy. A frozen frame-wise occupancy autoencoder maps occupancy to a compact latent space; for a sequence $O$, $\mathcal E_{\mathrm{occ}}(O)$ and $\mathcal D_{\mathrm{occ}}(z_o)$ denote applying the encoder or decoder independently at every horizon and stacking the resulting frames.

As illustrated in Fig.~\ref{fig:model_architecture}, \method combines an occupancy-native world--action generator with scene-to-ego two-stage flow matching. The generator uses a three-expert Mixture-of-Transformers backbone comprising Understanding, World, and Action Experts. The frozen Understanding Expert supplies semantic driving context, while the trainable World and Action Experts receive the current occupancy and trajectory flow states and exchange their hidden representations at every transformer layer. The two stages reuse the same expert backbone but employ representation-specific occupancy and action velocity heads.

Stage I transports independent Gaussian sources toward scene-centric occupancy and an absolute ego trajectory in a fixed frame anchored at the current ego pose. The resulting trajectory defines the future ego frames used to warp the decoded occupancy and is converted into adjacent pose increments. This transformed pair becomes the structured source of Stage II. 
The Stage-II occupancy target is instead the dataset's native future occupancy expressed in the ground-truth ego frame at every future horizon. Stage II then transports a model-induced source containing occupancy prediction errors, warp-induced uncovered regions, and pose misalignment toward a complete ego-centric world--action target without introducing new Gaussian noise.

Throughout, $r\in\{S,E\}$ indexes the scene-centric and ego-centric flow representations, respectively. In the $S$ representation, occupancy and cumulative ego poses are expressed in a fixed current-ego-anchored scene frame. In the $E$ representation, each occupancy slice is expressed in its corresponding future ego frame, while each action element is the adjacent pose increment, whose translational components are expressed along the fixed axes of a canonical sensor-aligned chart. We reserve $t\in[0,1]$ for flow time and $k\in\{1,\ldots,T\}$ for the physical future horizon. The symbol $\star$ denotes supervision targets, $\tilde{\cdot}$ the Stage-I output, $\bar{\cdot}$ the transformed Stage-II source, and $\hat{\cdot}$ the final Stage-II output. The operator $\sg(\cdot)$ denotes stop-gradient.

\subsection{Occupancy-Native World--Action Model}
\label{sec:occupancy_wam}

For representation $r\in\{S,E\}$, let $\mathbf{s}^{r}=(s_o^{r},s_a^{r})$ denote the source and let $\mathbf{x}^{r,\star}=(z_o^{r,\star},a^{r,\star})$ denote the target world--action state. We follow the standard source-to-target convention in which $t=0$ denotes the source and $t=1$ denotes the target. Both stages use the same shifted flow-time schedule. We first draw a uniform variable and transform it into the flow time as
\begin{equation}
    \tau\sim\mathcal U(0,1),
    \quad
    t
    =
    \gamma_{\rho}(\tau)
    =
    \frac{\tau}
    {\rho-(\rho-1)\tau},
    \quad
    \rho=5.
\label{eq:shifted_flow_time}
\end{equation}
The shift preserves the two endpoints while allocating more training samples near the source. We set the shift factor to 5 to strengthen learning in the early transport region, enabling accurate generation with fewer inference steps.

Given a paired source and target, the analytic conditional path used for training is
\begin{equation}
    \mathbf{x}_t^{r}
    =
    (1-t)\mathbf{s}^{r}
    +
    t\mathbf{x}^{r,\star},
    \quad
    \mathbf{u}^{r,\star}
    =
    \frac{\mathrm d\mathbf{x}_t^{r}}
         {\mathrm dt}
    =
    \mathbf{x}^{r,\star}
    -
    \mathbf{s}^{r}.
\label{eq:joint_flow_path}
\end{equation}
Here, $\mathbf{x}_t^{r}=(z_{o,t}^{r},a_t^{r})$ is the intermediate world--action state and $\mathbf{u}^{r,\star}=(u_o^{r,\star},u_a^{r,\star})$ is its exact conditional velocity. The World and Action streams share the same $t$ within each stage, while the two stages sample their flow times independently.

Let $H_l^u$, $H_l^{w,r}$, and $H_l^{a,r}$ denote the Understanding, World, and Action hidden representations at transformer layer $l\in\{0,\ldots,L-1\}$. Their asymmetric layer-wise interaction is
\begin{equation}
\begin{aligned}
    H_{l+1}^{u}
    &=
    U_l
    \left(
        H_l^{u}
    \right),\\
    H_{l+1}^{w,r}
    &=
    W_l
    \left(
        H_l^{w,r};
        H_l^{a,r},
        H_l^{u}
    \right),\\
    H_{l+1}^{a,r}
    &=
    A_l
    \left(
        H_l^{a,r};
        H_l^{w,r},
        H_l^{u}
    \right).
\end{aligned}
\label{eq:expert_interaction}
\end{equation}
The Understanding stream does not read the generated world or action states. In contrast, the World and Action Experts read both the semantic context and one another at every layer. A flow-time embedding is supplied to both generated streams before the shared expert backbone.

The expert blocks $\{U_l,W_l,A_l\}_{l=0}^{L-1}$ are reused across the two stages. The Understanding Expert remains frozen, whereas the World and Action Experts form the shared trainable backbone. Because the two stages use different occupancy and action representations, they employ four representation-specific velocity heads:
\begin{equation}
\begin{aligned}
    v_{o,\theta}^{r}
    \left(
        \mathbf{x}_t^{r},
        t;
        c
    \right)
    &=
    \Psi_o^{r}
    \left(
        H_L^{w,r}
    \right),\\
    v_{a,\theta}^{r}
    \left(
        \mathbf{x}_t^{r},
        t;
        c
    \right)
    &=
    \Psi_a^{r}
    \left(
        H_L^{a,r}
    \right),
    \qquad
    r\in\{S,E\}.
\end{aligned}
\label{eq:velocity_heads}
\end{equation}
The core world--action generator is parameterized by the shared World--Action backbone $\theta_{\mathrm{MoT}}$ and the four readouts $\{\Psi_o^S,\Psi_a^S,\Psi_o^E,\Psi_a^E\}$. We write the coupled vector field as $\mathbf{v}_{\theta}^{r}=(v_{o,\theta}^{r},v_{a,\theta}^{r})$. Although the final velocity components are read out separately, each head receives a hidden representation that has repeatedly interacted with the other generated stream. For $r=E$, the conditioning $c$ additionally includes the coverage pair defined below; we keep it implicit in $c$ to avoid overloading the notation.

Following conditional flow matching and rectified flow~\citep{lipman2023flow,liu2023rectified}, the objective for representation $r$ is
\begin{equation}
\label{eq:stage_flow_loss}
\begin{aligned}
    \mathcal L_{\mathrm{FM}}^{r}
    =
    \E_{\tau}
    \Big[
        &\lambda_o^{r}
        \bigl\|
            v_{o,\theta}^{r}(\mathbf{x}_t^{r},t;c)
            -
            u_o^{r,\star}
        \bigr\|_2^2\\
        +\;&
        \lambda_a^{r}
        \bigl\|
            v_{a,\theta}^{r}(\mathbf{x}_t^{r},t;c)
            -
            u_a^{r,\star}
        \bigr\|_2^2
    \Big],
\end{aligned}
\end{equation}
The expectation is taken over the training sample, its stage-specific source--target coupling, and $\tau\sim\mathcal U(0,1)$. The representation-specific weights balance the occupancy and action objectives under the two different parameterizations. For $r=E$, the occupancy flow term is supervised wherever valid native future-ego occupancy targets are available, including regions that are not covered by the transformed Stage-I source. The warp-coverage mask $M^E$ and token-wise coverage $\mu$ describe the \emph{source} and are used for conditioning, not to suppress the Stage-II loss. Uncovered source cells use a free-space placeholder only to form a categorical input; this placeholder is not treated as ground-truth free-space supervision.

At inference, we distinguish the numerical solver state $\mathbf{X}_n^{r}$ from the analytic interpolation $\mathbf{x}_t^{r}$ used for flow-matching supervision. For an $N$-step solver, we discretize the shifted time schedule as
\begin{equation}
    \tau_n
    =
    \frac{n}{N},
    \qquad
    t_n
    =
    \gamma_{\rho}(\tau_n),
    \qquad
    n=0,\ldots,N.
\label{eq:solver_time_grid}
\end{equation}
Starting from $\mathbf{X}_0^{r}=\mathbf{s}^{r}$, for $n=0,\ldots,N-1$ we apply the Euler update
\begin{equation}
    \mathbf{X}_{n+1}^{r}
    =
    \mathbf{X}_n^{r}
    +
    (t_{n+1}-t_n)\,
    \mathbf{v}_{\theta}^{r}
    (\mathbf{X}_n^{r},t_n;c).
\label{eq:shifted_euler_solver}
\end{equation}
The intermediate numerical states $\mathbf{X}_n^{r}$ are not assumed to coincide with the analytic interpolation states $\mathbf{x}_{t_n}^{r}$.

\subsection{Scene-to-Ego World--Action Flow Matching}
\label{sec:scene_to_ego_flow}

\paragraph{Stage I: Scene-centric joint flow matching.}
Let $\mathcal F^S$ denote a fixed scene frame anchored at the current ego pose and held constant over the prediction horizon. The scene-centric target consists of future occupancy $O^{S,\star}$ and an absolute ego trajectory $a^{S,\star}$, where \emph{absolute} means that all future poses are represented cumulatively in the same current-ego-anchored frame rather than in global-map coordinates. We encode the occupancy target and form the paired Stage-I target as
\begin{equation}
    z_o^{S,\star}
    =
    \mathcal E_{\mathrm{occ}}
    \left(
        O^{S,\star}
    \right),
    \qquad
    \mathbf{x}^{S,\star}
    =
    \left(
        z_o^{S,\star},
        a^{S,\star}
    \right).
\label{eq:stage_one_target}
\end{equation}
The Stage-I source consists of independent dimension-matched Gaussian variables:
\begin{equation}
    \mathbf{s}^{S}
    =
    (\epsilon_o^{S},\epsilon_a^{S}),
    \qquad
    \epsilon_m^{S}
    \sim
    \Normal(0,I_m),
    \quad m\in\{o,a\}.
\label{eq:stage_one_source}
\end{equation}
Stage I applies Eqs.~\eqref{eq:shifted_flow_time}--\eqref{eq:stage_flow_loss} with $r=S$.

We use $N_{\mathrm I}=1$ solver step for Stage I. Its scene-centric world--action output is
\begin{equation}
    \tilde{\mathbf{x}}^{S}
    :=
    \mathbf{X}_{N_{\mathrm I}}^{S}
    =
    \left(
        \tilde z_o^{S},
        \tilde a^{S}
    \right).
\label{eq:stage_one_output}
\end{equation}
The corresponding occupancy is decoded as $\tilde O^{S}=\mathcal D_{\mathrm{occ}}(\tilde z_o^{S})$. The trajectory $\tilde a^{S}=\{(\tilde{\mathbf p}_k,\tilde\psi_k)\}_{k=1}^{T}$ contains cumulative planar poses in $\mathcal F^S$. Stage I is kept to one solver evaluation because its output serves as a structured scene-centric world--action hypothesis and defines the future ego frames used by Stage II. The final occupancy completion and ego-centric transport are performed by the second flow.

\paragraph{Scene-to-ego transformation.}
Let $\Pi$ denote the fixed planar chart change that expresses a cumulative ego trajectory in the canonical sensor-aligned frame used by the Action Expert; it is a constant rigid transformation determined by the sensor extrinsics and does not depend on the trajectory. Writing $\Pi(a)=\{(\mathbf p_k,\psi_k)\}_{k=1}^{T}$ with $\mathbf p_0=\mathbf 0$ and $\psi_0=0$, the adjacent pose increments are
\begin{equation}
    [\Delta(a)]_k
    =
    \bigl(
        \mathbf p_k-\mathbf p_{k-1},\;
        \operatorname{wrap}(\psi_k-\psi_{k-1})
    \bigr),
\label{eq:ego_relative_action}
\end{equation}
for $k=1,\ldots,T$, where $\operatorname{wrap}(\cdot)$ maps an angle to $(-\pi,\pi]$.
The translational increments are differences taken along the fixed axes of the canonical chart rather than in the instantaneous body frame. The inverse operator $\Delta^{-1}$ accumulates the increments component-wise and applies $\Pi^{-1}$ to recover a cumulative trajectory in $\mathcal F^S$.

Let $\mathcal W(O,a)$ re-index every scene-centric occupancy slice into the future ego frame induced by the corresponding cumulative pose in $a$, returning the re-indexed occupancy together with a binary validity mask. Only the detached Stage-I
prediction is geometrically transformed:
\begin{equation}
    \bigl(\bar O^{E},\,M^{E}\bigr)
    =
    \mathcal W
    \bigl(
        \sg(\tilde O^{S}),\;
        \sg(\tilde a^{S})
    \bigr).
\label{eq:stage_two_warp}
\end{equation}
For every destination cell, $\mathcal W$ inverse-samples the corresponding location in the scene-centric prediction using nearest-neighbour sampling, and the binary mask $M^{E}$ records whether that location falls inside the valid scene-centric extent.
Because the future ego frames advance along the predicted trajectory, part of every destination grid necessarily lies beyond that extent, so $\bar O^{E}$ contains uncovered regions that grow with the prediction horizon. Such cells are assigned the
free-space class solely to provide the frozen autoencoder with a valid categorical input; $M^{E}$ is evaluated before this filling, so their identity as uncovered is carried by the mask rather than inferred from the filled value.

Averaging $M^{E}$ over the voxel footprint of each latent token yields a visible
fraction $\mu\in[0,1]$, and the pair $[\mu,\,1-\mu]$ is injected twice: before the World Expert, so
that occupancy-flow tokens carry their own coverage, and again before the occupancy
readout. The Stage-II source is
{\small
\begin{equation}
    \bar z_o^{E}
    =
    \mathcal E_{\mathrm{occ}}(\bar O^{E}),
    \quad
    \bar a^{E}
    =
    \Delta\bigl(\sg(\tilde a^{S})\bigr),
    \quad
    \mathbf{s}^{E}
    =
    (\bar z_o^{E},\bar a^{E}).
\label{eq:stage_two_source_components}
\end{equation}
}
with $\mu$ supplied as an additional deterministic condition alongside $c$.

\paragraph{Native future-ego target.}
The Stage-II occupancy target
$O^{E,\star}=\{O_k^{E,\star}\}_{k=1}^{T}$
is the dataset's native future occupancy expressed in the ground-truth ego frame at
each horizon. It is not obtained by warping $O^{S,\star}$ and therefore carries real
content in the regions left uncovered when constructing $\bar O^{E}$. The paired
target is $\mathbf{x}^{E,\star}=(z_o^{E,\star},a^{E,\star})$ with
\begin{equation}
    z_o^{E,\star}
    =
    \mathcal E_{\mathrm{occ}}(O^{E,\star}),
    \qquad
    a^{E,\star}
    =
    \Delta(a^{S,\star}).
\label{eq:stage_two_target_components}
\end{equation}
No ground-truth future pose is used to transform a model prediction; ground-truth ego
poses enter only through the native occupancy supervision and the action target.
Here \emph{native} specifies that supervision is defined in each ground-truth future ego frame rather than produced by warping the Stage-I scene-centric target. It does not denote an inference-time input or require ground-truth future occupancy to be provided to the deployed policy. The Stage-II warp-coverage mask tracks source support, not target-label availability.

Source and target share the same canonical layouts but are induced by different pose
sequences: the source occupancy follows the future ego frames predicted by Stage I,
whereas the target occupancy follows the ground-truth ones. Stage-I pose error
therefore appears simultaneously as spatial misalignment in the occupancy stream and
as error in the pose increments, while the source additionally carries occupancy
prediction errors and uncovered regions. These discrepancies share a single origin,
and the World and Action Experts transport them jointly toward the paired ego-centric
target.

\begin{table*}[t]
\centering
\caption{Closed-loop driving performance on Bench2Drive.
\emph{Expert} denotes the expert policy used for collecting training data,
while \emph{WM} denotes an explicit learned future-world model.
Best and second-best results are shown in bold and underlined, respectively.}
\label{tab:b2d}

\small
\setlength{\tabcolsep}{4.2pt}
\renewcommand{\arraystretch}{1.04}

\begin{tabular*}{0.98\textwidth}
{@{\extracolsep{\fill}}lccccc@{}}
\toprule
Method
& Venue
& Expert
& WM
& DS $\uparrow$
& SR (\%) $\uparrow$ \\
\midrule

UniAD-Base~\citep{hu2023uniad}
& CVPR'23
& Think2Drive
& \xmark
& 45.81
& 16.36 \\

DiffusionDrive~\citep{liao2025diffusiondrive}
& CVPR'25
& PDM-Lite
& \xmark
& 77.68
& 57.72 \\

ORION~\citep{fu2025orion}
& ICCV'25
& Think2Drive
& \xmark
& 77.74
& 54.62 \\

AutoVLA~\citep{zhou2025autovla}
& NeurIPS'25
& PDM-Lite
& \xmark
& 78.84
& 57.73 \\

TransFuser++~\citep{jaeger2023hidden}
& ICCV'23
& PDM-Lite
& \xmark
& 84.21
& 67.27 \\

SimLingo~\citep{renz2025simlingo}
& CVPR'25
& PDM-Lite
& \xmark
& 85.07
& 67.27 \\

AutoMoT+~\citep{huang2026automot}
& ICML'26
& PDM-Lite
& \xmark
& \underline{89.42}
& \underline{74.09} \\

\midrule

WoTE~\citep{li2025wote}
& ICCV'25
& Think2Drive
& \cmark
& 61.71
& 31.36 \\

UniDrive-WM~\citep{xiong2026unidrivewm}
& ECCV'26
& Think2Drive
& \cmark
& 79.31
& 56.42 \\

DeepSight~\citep{zhang2026deepsight}
& ICML'26
& Think2Drive
& \cmark
& 86.23
& 71.36 \\

KnowVal-SimLingo~\citep{xia2026knowval}
& CVPR'26
& PDM-Lite
& \cmark
& 88.42
& 69.03 \\

\midrule

\method
& --
& PDM-Lite
& \cmark
& \textbf{90.45}
& \textbf{77.27} \\

\bottomrule
\end{tabular*}
\end{table*}

\paragraph{Stage II: Ego-centric joint flow matching.}
Stage II applies Eqs.~\eqref{eq:shifted_flow_time}--\eqref{eq:stage_flow_loss} with
source $\mathbf{s}^{E}$ and target $\mathbf{x}^{E,\star}$. It introduces no independent Gaussian source. Although the source distribution is model-induced and generally non-Gaussian, the paired Stage-I output and dataset supervision define a valid
conditional source--target coupling under the same affine flow path.

Using $N_{\mathrm{II}}$ Euler steps, Stage II returns
\begin{equation}
    \hat{\mathbf{x}}^{E}
    :=
    \mathbf{X}_{N_{\mathrm{II}}}^{E}
    =
    \left(
        \hat z_o^{E},
        \hat a^{E}
    \right).
\label{eq:stage_two_output}
\end{equation}
The final occupancy is decoded as $\hat O^{E}=\mathcal D_{\mathrm{occ}}(\hat z_o^{E})$. It follows the dataset's native
ego-centric output layout, and no additional geometric warp is applied after Stage II. The cumulative trajectory is recovered as $\hat a^{S}=\Delta^{-1}(\hat a^{E})$.

\paragraph{Training and inference.}
The occupancy autoencoder and Understanding Expert remain frozen during world--action training. Stage I is trained from independent Gaussian sources through $\mathcal L_{\mathrm{FM}}^{S}$. During every training iteration, its one-step output is detached, decoded, transformed through $\mathcal W$, and re-encoded to construct the Stage-II source. Stage II independently samples $\tau$, uses the same shifted time map, and is trained through $\mathcal L_{\mathrm{FM}}^{E}$. 

Beyond the two flow-matching objectives, we introduce an auxiliary
world--action coupling loss that evaluates the predicted trajectory
against the predicted occupancy field.
Let $\Phi(P,a)$ denote an occupancy-based trajectory cost that samples
the per-class probability field $P$ along the ego footprint induced
by trajectory $a$.

In Stage I, the cost is evaluated in the fixed scene-centric frame,
penalizing departure from drivable regions and overlap with static
and dynamic obstacles.
In Stage II, the adjacent pose increments are accumulated to recover
the trajectory in the fixed current-ego coordinate frame.
For coupling-cost evaluation, the occupancy probability field and
the reconstructed trajectory are expressed in the same coordinate
frame, and the same penalties are applied.

Both coupling terms are evaluated as
$\Phi(\operatorname{sg}(P),a)$, where $\operatorname{sg}$
denotes stop-gradient.
This prevents the coupling loss from directly back-propagating
through the predicted occupancy probabilities, avoiding the
degenerate solution of reducing collision penalties by removing
obstacles from the predicted world.

The common world--action objective is
\begin{equation}
    \mathcal{L}
    =
    \mathcal{L}_{\mathrm{FM}}^{S}
    +
    \mathcal{L}_{\mathrm{couple}}^{S}
    +
    \lambda_{\mathrm{II}}
    \bigl(
        \mathcal{L}_{\mathrm{FM}}^{E}
        +
        \mathcal{L}_{\mathrm{couple}}^{E}
    \bigr).
\label{eq:overall_training_objective}
\end{equation}

Additional benchmark-specific auxiliary objectives are incorporated
during training, as detailed in Appendix B.2.

The Stage-II loss does not back-propagate through the Stage-I solver output or the geometric warp. Both losses nevertheless update the shared World and Action Expert backbone through their respective invocations, while each stage updates its own occupancy and action velocity heads.

At inference, the frozen Understanding Expert is evaluated once, and its
context is reused across both stages. Each stage uses a single World--Action backbone pass for all reported results.

\begin{table*}[!t]
\centering
\caption{Open-loop planning and future-occupancy evaluation on nuScenes.}
\label{tab:nusc}
\footnotesize

\textbf{(a) Open-loop planning.}\par\smallskip

{
\setlength{\tabcolsep}{2.7pt}
\renewcommand{\arraystretch}{1.04}
\begin{tabular*}{\textwidth}{@{\extracolsep{\fill}}llcccccccc@{}}
\toprule
& & \multicolumn{4}{c}{L2 (m) $\downarrow$} & \multicolumn{4}{c}{Collision (\%) $\downarrow$} \\
\cmidrule(lr){3-6}\cmidrule(lr){7-10}
Method & Ego status & 1s & 2s & 3s & Avg. & 1s & 2s & 3s & Avg. \\
\midrule
UniAD~\citep{hu2023uniad} & Vector & 0.44 & 0.67 & 0.96 & 0.69 & 0.04 & 0.08 & 0.23 & 0.12 \\
VAD~\citep{jiang2023vad} & Vector & 0.17 & 0.34 & 0.60 & 0.37 & 0.07 & 0.10 & 0.24 & 0.14 \\
DriveTransformer-Large~\citep{jia2025drivetransformer} & Vector & 0.16 & 0.30 & 0.55 & 0.33 & \underline{0.01} & \underline{0.06} & \underline{0.15} & \textbf{0.07} \\
AutoVLA~\citep{zhou2025autovla} & Text & 0.21 & 0.38 & 0.60 & 0.40 & 0.13 & 0.18 & 0.28 & 0.20 \\
OmniDrive~\citep{wang2025omnidrive} & Vector & \textbf{0.14} & \underline{0.29} & 0.55 & 0.33 & \textbf{0.00} & 0.13 & 0.78 & 0.30 \\
SpaceDrive~\citep{li2026spacedrive} & Vector & \underline{0.15} & \underline{0.29} & 0.51 & 0.32 & 0.04 & 0.18 & 0.49 & 0.23 \\
AutoMoT~\citep{huang2026automot} & Vector & \textbf{0.14} & \underline{0.29} & 0.54 & 0.32 & \underline{0.01} & \underline{0.06} & \underline{0.15} & \textbf{0.07} \\
DriveVLM-Dual~\citep{tian2025drivevlm} & Vector & \underline{0.15} & \underline{0.29} & \textbf{0.48} & \underline{0.31} & 0.05 & 0.08 & 0.17 & \underline{0.10} \\
\midrule
\method & Vector & \textbf{0.14} & \textbf{0.27} & \underline{0.50} & \textbf{0.30} & 0.02 & \textbf{0.05} & \textbf{0.14} & \textbf{0.07} \\
\bottomrule
\end{tabular*}
}

\medskip
\textbf{(b) Future semantic occupancy.}\par\smallskip

{
\setlength{\tabcolsep}{2.8pt}
\renewcommand{\arraystretch}{1.04}
\begin{tabular*}{\textwidth}{@{\extracolsep{\fill}}lllcccccccc@{}}
\toprule
& & & \multicolumn{4}{c}{mIoU (\%) $\uparrow$} & \multicolumn{4}{c}{IoU (\%) $\uparrow$} \\
\cmidrule(lr){4-7}\cmidrule(lr){8-11}
Method & Input & Ego traj. & 1s & 2s & 3s & Avg. & 1s & 2s & 3s & Avg. \\
\midrule
Copy \& Paste~\citep{zheng2024occworld} & 3D-Occ & Pred. & 14.91 & 10.54 & 8.52 & 11.33 & 24.47 & 19.77 & 17.31 & 20.52 \\
OccWorld~\citep{zheng2024occworld} & 3D-Occ & Pred. & 25.78 & 15.14 & 10.51 & 17.14 & 34.63 & 25.07 & 20.18 & 26.63 \\
OccLLaMA~\citep{wei2024occllama} & 3D-Occ & Pred. & 25.05 & 19.49 & 15.26 & 19.93 & 34.56 & 28.53 & 24.41 & 29.17 \\
OccTENS~\citep{jin2025occtens} & 3D-Occ & Pred. & 27.96 & \underline{21.75} & \textbf{16.47} & 22.06 & 38.73 & 29.50 & 24.86 & 31.03 \\
RenderWorld~\citep{yan2025renderworld} & 3D-Occ & Pred. & 28.69 & 18.89 & 14.83 & 20.80 & 37.74 & 28.41 & 24.08 & 30.08 \\
DFIT-OccWorld~\citep{zhang2024dfitoccworld} & 3D-Occ & Pred. & \underline{31.68} & 21.29 & 15.18 & \underline{22.71} & \underline{40.28} & \underline{31.24} & \underline{25.29} & \underline{32.27} \\
COME~\citep{shi2025come} & 3D-Occ & Pred. & 30.57 & 19.91 & 13.38 & 21.29 & 36.96 & 28.26 & 21.86 & 29.03 \\
\midrule
\method & 3D-Occ & Pred. & \textbf{34.02} & \textbf{23.12} & \underline{15.55} & \textbf{24.23} & \textbf{43.76} & \textbf{32.30} & \textbf{25.57} & \textbf{33.88} \\
\bottomrule
\end{tabular*}
}
\end{table*}

\section{Experiments}
\label{sec:experiments}

\subsection{Experimental Setup}
\label{sec:experimental_setup}

\paragraph{Benchmarks and metrics.}
Bench2Drive is our primary closed-loop benchmark, where we report Driving Score (DS) and Success Rate (SR)~\citep{jia2024bench2drive}. On nuScenes~\citep{caesar2020nuscenes}, we evaluate open-loop planning using average L2 displacement error and collision rate. We additionally evaluate future volumetric occupancy on Occ3D using semantic mIoU and geometric IoU~\citep{hu2022stp3,tian2023occ3d}. All results follow the official validation splits and evaluation protocols. For nuScenes, open-loop planning uses camera observations, whereas future-occupancy prediction uses four historical 3D occupancy frames; these are task-specific evaluation configurations, not outputs of one identical raw-input forward pass (Appendix~\ref{sec:app_benchmark_modules}). Bench2Drive uses front RGB and LiDAR BEV observations at inference and does not use ground-truth occupancy as a runtime input.

\paragraph{Understanding-Expert adaptation.}
Before world--action training, we adapt Qwen3-VL-2B-Instruct to driving visual question answering using LoRA. The Bench2Drive expert receives the current front-camera frame together with three history frames and is trained on $56{,}004$ Chat-B2D front-view question--answer examples~\citep{fu2025orion}. The nuScenes expert is trained on $50{,}008$ examples from nuScenes-QA and OmniDrive~\citep{qian2024nuscenesqa,wang2025omnidrive}. The adapters are merged into benchmark-specific checkpoints and remain frozen throughout world--action training.

\subsection{Closed-Loop Driving on Bench2Drive}
\label{sec:b2d_results}

We first evaluate \method on the Bench2Drive closed-loop benchmark and report
the quantitative results in Table~\ref{tab:b2d}. \method achieves the best
performance in both Driving Score and Success Rate, reaching $90.45$ and
$77.27\%$, respectively. It surpasses AutoMoT+~\citep{huang2026automot}, the
strongest expert-only baseline with diffusion-based action refinement (DS
$89.42$, SR $74.09\%$), and KnowVal-SimLingo~\citep{xia2026knowval}, the
strongest prior world-model-based system (DS $88.42$, SR $69.03\%$).

The improvement highlights the benefit of the occupancy-native
world--action formulation. By representing future occupancy and ego motion
within a shared generative state, OccWAM provides planning with an explicit,
metric description of the evolving scene. This unified representation
preserves the spatial structure of the predicted future throughout action
generation, providing a structured basis for accurate and safe planning.

\subsection{Open-Loop Planning and Future Occupancy on nuScenes}
\label{sec:nusc_results}

Table~\ref{tab:nusc} evaluates the action and world components under a fixed driving log. The planning metrics measure trajectory imitation and geometric collision, while the occupancy metrics verify that the World Expert produces an independently meaningful metric future rather than an unconstrained hidden feature optimized only for action prediction. \method achieves the best average planning L2 and collision rate, together with the best average mIoU and IoU.

The two evaluations are reported together to examine both sides of the generated world--action state. Strong planning without a measurable future world would reduce occupancy to an opaque internal representation, whereas accurate occupancy without a planning gain would not establish an action-facing predictive state. The results show that \method improves planning while preserving a high-quality explicit future world.

\subsection{Ablation Studies}
\label{sec:ablations}

We investigate two key components of \method:
the scene-to-ego two-stage joint flow and bidirectional
information exchange between the World and Action Experts.
We first examine their effects on planning performance,
followed by complementary analyses of Action-to-World
information flow and joint world--action refinement on nuScenes.

\begin{table}[!htbp]
\centering
\caption{Ablation of the second joint flow and World-to-Action information flow.}
\label{tab:core_ablations}
\footnotesize
\begingroup
\setlength{\tabcolsep}{3pt}
\renewcommand{\arraystretch}{1.12}

\textbf{(a) Contribution of two-stage flow on Bench2Drive.}
\par\smallskip
\begin{tabular*}{\columnwidth}{@{\extracolsep{\fill}}lcc@{}}
\toprule
Control output & DS $\uparrow$ & SR (\%) $\uparrow$ \\
\midrule
Stage-I trajectory & 88.21 & 73.63 \\
Stage-II trajectory & \textbf{90.45} & \textbf{77.27} \\
\bottomrule
\end{tabular*}

\medskip
\textbf{(b) Effect of World-to-Action information flow.}
\par\smallskip
\begin{tabular*}{\columnwidth}{@{\extracolsep{\fill}}lcc@{}}
\toprule
Configuration & L2 (m) $\downarrow$ & Collision (\%) $\downarrow$ \\
\midrule
Without World information flow & 0.64 & 0.23 \\
Full \method & \textbf{0.30} & \textbf{0.07} \\
\bottomrule
\end{tabular*}
\endgroup
\end{table}

\paragraph{Effect of the two-stage joint flow.}
Using the final Stage-II trajectory improves DS from
$88.21$ to $90.45$ and SR from $73.63\%$ to $77.27\%$
on Bench2Drive. This gain is consistent with the
complementary roles of the two stages.
Stage I establishes a globally coherent scene-centric
world--action hypothesis, whereas Stage II re-expresses
the joint future in trajectory-induced ego coordinates,
bringing the predicted world into closer alignment
with the geometry relevant to execution.
The second flow thus bridges the representation gap
between scene-level future modeling and ego-centric
action generation, enabling more action-aligned reasoning
over both occupancy and motion.
These results support the effectiveness of the proposed
scene-to-ego decomposition for closed-loop planning.

\paragraph{Effect of World-to-Action information flow.}
We disable the Action Expert's access to the World Expert
during inference while keeping the trained model and all
other inputs unchanged.
Removing this information pathway increases the planning
L2 error from $0.30$ to $0.64$\,m and the collision rate
from $0.07\%$ to $0.23\%$ on nuScenes.
The substantial degradation indicates that access to
the evolving occupancy representation is important
for trajectory generation.
In particular, the larger relative increase in collision
rate suggests that the predicted future world provides
spatial information relevant to safe motion.
This inference-time intervention demonstrates the
functional importance of World-to-Action information
exchange in the joint generation process.

\begin{table}[!htbp]
\centering
\caption{
Complementary ablations on nuScenes.
Planning metrics (L2 and collision) use camera observations,
whereas occupancy metrics (mIoU and IoU) use four historical
3D occupancy frames. Comparisons within each metric group
follow the same task-specific input and evaluation protocol.
}
\label{tab:additional_ablations}
\footnotesize
\begingroup
\setlength{\tabcolsep}{2.5pt}
\renewcommand{\arraystretch}{1.12}

\textbf{(c) Effect of Action-to-World information flow.}
\par\smallskip
\begin{tabular*}{\columnwidth}{@{\extracolsep{\fill}}lcccc@{}}
\toprule
Configuration
& L2 $\downarrow$
& Coll. $\downarrow$
& mIoU $\uparrow$
& IoU $\uparrow$ \\
\midrule
Without Action-to-World
& 0.32 & 0.10 & 23.11 & 32.59 \\
Full \method
& \textbf{0.30}
& \textbf{0.07}
& \textbf{24.23}
& \textbf{33.88} \\
\bottomrule
\end{tabular*}

\medskip
\textbf{(d) Source-to-output refinement in Stage II.}
\par\smallskip
\begin{tabular*}{\columnwidth}{@{\extracolsep{\fill}}lcccc@{}}
\toprule
Readout
& L2 $\downarrow$
& Coll. $\downarrow$
& mIoU $\uparrow$
& IoU $\uparrow$ \\
\midrule
Stage-II source
& 0.32 & 0.13 & 23.12 & 31.78 \\
Stage-II output
& \textbf{0.30}
& \textbf{0.07}
& \textbf{24.23}
& \textbf{33.88} \\
\bottomrule
\end{tabular*}

\endgroup
\end{table}

\paragraph{Effect of Action-to-World information flow.}
To complement the World-to-Action analysis, we examine
the reverse information pathway by preventing the
World Expert from attending to the Action Expert
during inference.
Within each task-specific nuScenes protocol, the trained
configuration, flow states, and evaluation procedure
remain unchanged, with the cross-expert information
pathway being the only intervention.

As shown in Table~\ref{tab:additional_ablations}(c),
disabling Action-to-World information flow increases
the planning L2 error from $0.30$ to $0.32$\,m and
the collision rate from $0.07\%$ to $0.10\%$.
Meanwhile, occupancy mIoU decreases from $24.23$
to $23.11$, and IoU decreases from $33.88$ to $32.59$.

These results indicate that the evolving action
representation also provides useful information
for future-world generation.
Together with the World-to-Action analysis, they
support the functional importance of bidirectional
interaction between the two experts.
Since the information pathway is disabled only
during inference, this experiment is interpreted
as an information-flow diagnostic rather than
a separately retrained architectural ablation.

\paragraph{Joint world--action refinement in Stage II.}
Beyond the closed-loop improvement on Bench2Drive,
we further examine whether the second joint flow
refines both components of the generated world--action
state on nuScenes.

We compare the transformed Stage-II source
$(\bar z_o^E,\bar a^E)$ with the final Stage-II output
$(\hat z_o^E,\hat a^E)$.
For planning evaluation, the source action
$\bar a^E$ is reconstructed from adjacent ego-pose
increments into an absolute trajectory and evaluated
under the camera-based planning protocol.
For occupancy evaluation, the source occupancy
latent $\bar z_o^E$ is decoded and evaluated against
the same native future-ego occupancy target used
for the final Stage-II output, under the four-frame
historical-occupancy protocol.
Within each metric group, the source and output
share identical inputs and evaluation settings.

Table~\ref{tab:additional_ablations}(d) shows that
the second joint flow reduces the planning L2 error
from $0.32$ to $0.30$\,m and the collision rate
from $0.13\%$ to $0.07\%$.
Concurrently, occupancy mIoU increases from $23.12$
to $24.23$, while IoU improves from $31.78$ to
$33.88$, corresponding to absolute gains of
$1.11$ and $2.10$ percentage points, respectively.

These improvements demonstrate that Stage II refines
both the trajectory and the ego-centric future occupancy,
rather than acting merely as an action-only
post-processing module.
By transforming the scene-centric joint hypothesis
into trajectory-induced ego coordinates, the second
flow operates on a representation aligned with
future ego motion, allowing the world and action
states to be refined jointly.
The occupancy improvement is also consistent with
the ability of Stage II to correct transformed
world predictions and infer content in regions
not covered by the Stage-I source.

Overall, these results support the complementary
roles of the two stages: Stage I constructs a
scene-centric joint future hypothesis, while
Stage II re-indexes and jointly refines the
world--action state in future ego-centric coordinates.
Combined with the bidirectional information-flow
analyses, they provide additional evidence for
the proposed joint scene-to-ego generation framework.

\section{Conclusion}
We presented \method, an occupancy-native world--action model that jointly
generates dense metric occupancy and continuous ego motion through coupled
flow matching. A shared three-expert Mixture-of-Transformers enables direct
World--Action interaction, making occupancy an explicit and measurable future
representation that participates in action generation. Our scene-to-ego design
first predicts scene-centric occupancy and an absolute trajectory, then
transports the transformed hypothesis toward an ego-centric occupancy--motion
endpoint through a second joint flow. Experiments on Bench2Drive and nuScenes
demonstrate state-of-the-art performance in closed-loop driving, open-loop
planning, and future-occupancy prediction.
\begingroup
\small
\setlength{\bibsep}{1pt}
\bibliographystyle{unsrtnat}
\bibliography{references}
\endgroup

\clearpage
\appendix
\setcounter{table}{0}
\renewcommand{\thetable}{S\arabic{table}}
\setcounter{figure}{0}
\renewcommand{\thefigure}{S\arabic{figure}}
\section*{Supplementary Material}

\newcommand{\OccWAMQualitativeOccFigure}{%
  \begin{figure*}[!t]
    \centering
    \includegraphics[width=0.94\textwidth,
      height=0.52\textheight,keepaspectratio]
      {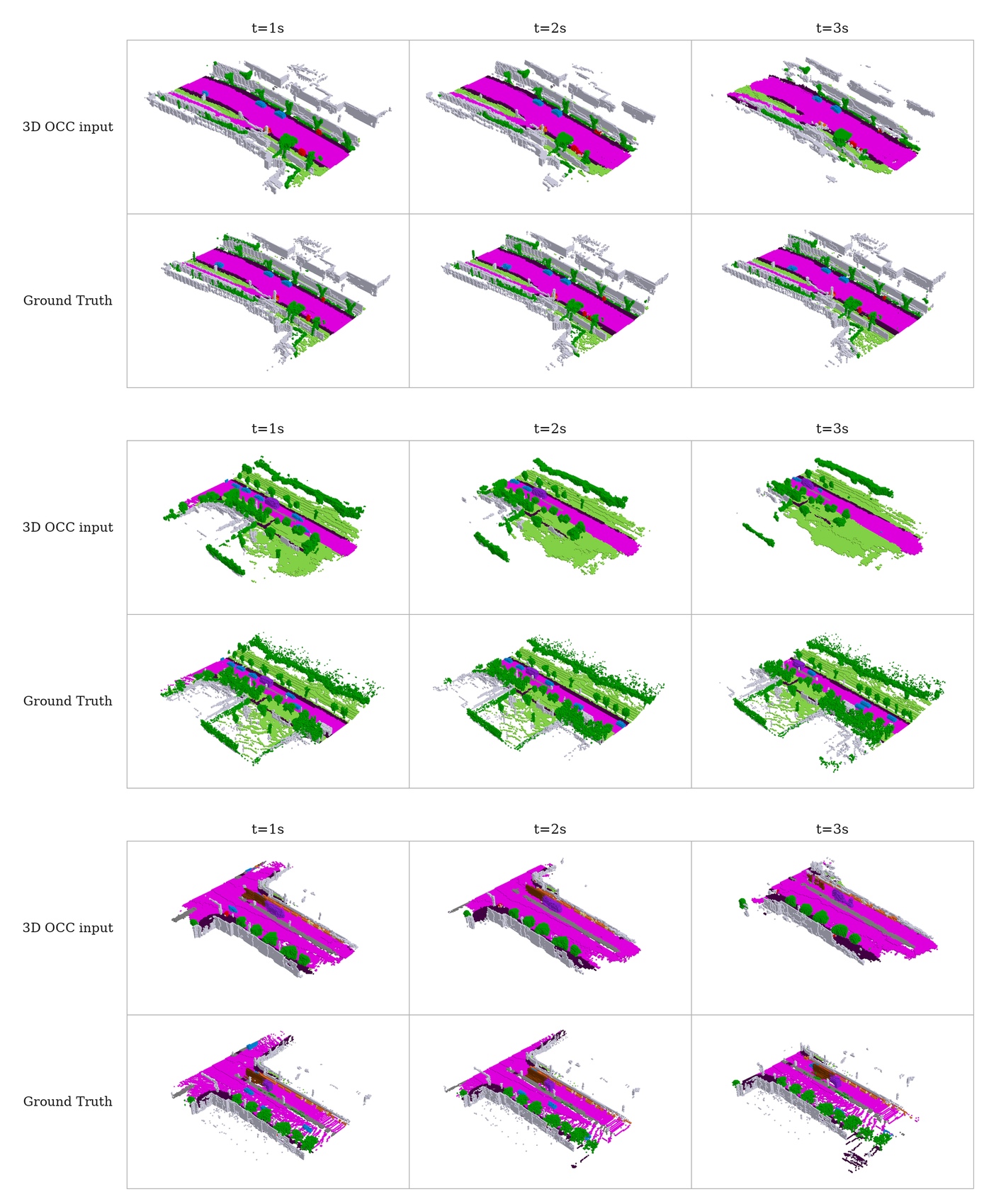}
    \caption{Qualitative comparison of future 4D occupancy predictions and
      ground truth across future time steps. Each frame shows the predicted
      3D semantic occupancy of \method and its corresponding ground truth.}
    \label{fig:app_4d_occupancy}
  \end{figure*}%
}
\begingroup
  \let\OccWAMOriginalLabel\label
  \def\OccWAMTargetLabel{sec:app_benchmark_modules}
  \renewcommand{\label}[1]{%
    \OccWAMOriginalLabel{#1}%
    \def\OccWAMCurrentLabel{#1}%
    \ifx\OccWAMCurrentLabel\OccWAMTargetLabel
      \OccWAMQualitativeOccFigure
    \fi
  }
  \section{Overview}

This appendix accompanies the main paper in the same PDF.
It details benchmark-specific occupancy modules, world-state
representations, and training procedures, including the
distinct nuScenes input protocols for planning and occupancy
prediction. We further clarify the coverage conditioning
and supervision used in Stage II, together with the auxiliary
world--action coupling objectives.

\section{Additional Implementation Details}

\subsection{Benchmark-Specific Occupancy Modules}
\label{sec:app_benchmark_modules}

We use the term benchmark-specific Occ Encoder to refer to the complete occupancy module associated with each benchmark. It connects the observations available under the corresponding task protocol to the World Expert and also contains the mappings between metric occupancy and latent occupancy states. Depending on the benchmark and evaluation task, its input-processing pathway converts sensor observations or historical occupancy states into a compact world representation supplied to the World Expert.

Within this module, $\mathcal E_{\mathrm{occ}}$ and $\mathcal D_{\mathrm{occ}}$ map semantic occupancy between the metric grid and the latent flow space. Specifically, $\mathcal E_{\mathrm{occ}}$ encodes semantic occupancy targets or transformed occupancy states into latent representations, while $\mathcal D_{\mathrm{occ}}$ decodes predicted occupancy latents back into the metric semantic grid.

Bench2Drive neither provides nor uses historical or current ground-truth occupancy as model input. Its Occ Encoder takes one front-view RGB image and one LiDAR BEV representation and produces the latent world features supplied to the World Expert. The model predicts six future BEV occupancy frames at $2$ Hz over a $64\,\mathrm{m}\times64\,\mathrm{m}$ ego-centered region with a spatial resolution of $1\,\mathrm{m}$. Each predicted frame is represented as a $64\times64$ semantic grid with $11$ categorical classes, including free space. In the latent flow space, the complete future occupancy sequence has shape $6\times8\times8\times8$ $(T\times H\times W\times C)$, corresponding to $384$ occupancy tokens of dimension $8$.

nuScenes uses two task-specific input protocols. For the planning benchmark, the model receives camera observations, and L2 and collision are evaluated under this camera-based planning protocol. For occupancy prediction, the benchmark-specific Occ Encoder jointly processes four historical 3D occupancy frames and compresses their temporal information into a compact set of $625$ history tokens. The four-frame history is therefore represented by $625$ temporally condensed tokens in total, rather than by four separate sets of $625$ tokens. Only these condensed history tokens are supplied to the World Expert, preventing the Mixture-of-Transformers from directly processing separate token sets for all four historical frames.

The nuScenes occupancy-prediction target consists of six future volumetric occupancy frames indexed by $(t,x,y,z)$. Each frame covers the spatial range $[-40,40]\times[-40,40]\times[-1,5.4]\,\mathrm{m}$ at a voxel resolution of $0.4\,\mathrm{m}$ and is decoded into a dense $200\times200\times16$ grid with $17$ semantic classes plus free space. The reported mIoU and IoU are computed under this four-frame historical-occupancy protocol.

Accordingly, planning and occupancy numbers shown in the same result row should be interpreted as evaluations of the corresponding task-specific configurations of the same method under their respective benchmark protocols, rather than as outputs produced by the same raw-input forward pass. Within each metric group, all compared configurations use the same input protocol and evaluation procedure.

The Understanding Expert follows the same training protocol on both benchmarks. It is first adapted using benchmark-specific question--answering supervision and is subsequently frozen throughout World--Action training.

\paragraph{Bench2Drive spatial route prediction.}
Bench2Drive additionally requires the prediction of $20$ spatial route points representing the reference-path geometry. These points are distinct from the temporal ego trajectory generated by the joint occupancy--action flows. We append a dedicated learnable route query after the occupancy and action tokens and decode its final representation directly into the route-point sequence.

The route query is supervised by the benchmark-provided route points and remains outside the joint flow. It receives no Gaussian source, flow-time interpolation, or velocity-matching objective. This benchmark-specific route branch is used only on Bench2Drive.

\subsection{Model and Training Configuration}

World--Action training optimizes the World and Action Experts, four representation-specific output decoders, and the benchmark-specific prediction modules. The four output decoders correspond to the occupancy and action outputs of the scene-centric and ego-centric flows. They implement the representation-specific velocity readouts in the main formulation and are distinct from the occupancy decoder $\mathcal D_{\mathrm{occ}}$ contained in the Occ Encoder.

The benchmark-specific trainable modules include the Bench2Drive route branch, the associated action and trajectory modules, newly introduced projection layers, and the scene-prior module used by the Bench2Drive configuration. The Understanding-Expert context is computed once and reused across both stages, and each stage requires one World--Action backbone pass.

Stage I and Stage II are optimized jointly rather than trained as separate models. In the notation of Eq.~\eqref{eq:overall_training_objective}, we set $\lambda_{\mathrm{II}}=0.5$, so the aggregate Stage-II objective receives one half of the weight assigned to its Stage-I counterpart. Equivalently, the Stage-I and Stage-II aggregate losses are weighted at a ratio of $2{:}1$. The same weighting is used for both Bench2Drive and nuScenes.

Bench2Drive uses two training phases. Phase 1 adapts the Understanding Expert using benchmark-specific question--answering supervision. Phase 2 freezes the Understanding Expert and jointly trains Stage I and Stage II with the weighting described above.

nuScenes uses three training phases. Phases 1 and 2 follow the same high-level procedure: question--answering adaptation is followed by joint Stage-I and Stage-II training with the same $2{:}1$ weighting. In Phase 3, all other components are frozen, and only the Stage-II occupancy and action output decoders are fine-tuned.

\paragraph{Benchmark-specific auxiliary objectives.}
Equation~\eqref{eq:overall_training_objective} specifies the joint flow-matching and World--Action coupling objective shared by both benchmarks. In addition to these common terms, Bench2Drive uses decoded occupancy reconstruction and trajectory-consistency objectives, while nuScenes uses an ego-future prediction objective.

The decoded occupancy reconstruction objective supervises semantic occupancy after the predicted occupancy latent has been passed through $\mathcal D_{\mathrm{occ}}$. This decoded semantic supervision is propagated to the World Expert and occupancy output decoder.

The trajectory-consistency objective provides additional supervision for the Bench2Drive trajectory-related modules. The ego-future prediction objective provides benchmark-specific supervision for the nuScenes future-ego representation. These auxiliary terms complement the latent flow supervision without changing the scene-to-ego formulation or the inference procedure.

Table~\ref{tab:supp_implementation} summarizes the principal benchmark-specific configurations.

\begin{table*}[t]
\centering
\caption{Benchmark-specific model and training configurations. The planning and occupancy-prediction rows distinguish the task-specific nuScenes input protocols.}
\label{tab:supp_implementation}
\footnotesize
\setlength{\tabcolsep}{4.0pt}
\renewcommand{\arraystretch}{1.10}
\begin{tabularx}{\textwidth}{
    @{}
    >{\raggedright\arraybackslash}p{0.19\textwidth}
    Y
    Y
    @{}
}
\toprule
Setting & Bench2Drive & nuScenes \\
\midrule

Planning input
& Front-view RGB image and one LiDAR BEV representation
& Camera observations \\

Occupancy-prediction input
& The same front-view RGB and LiDAR BEV observations; no historical or current ground-truth occupancy is used
& Four historical 3D occupancy frames \\

Predicted occupancy
& Temporal BEV occupancy over $(t,x,y)$, without an explicit vertical dimension
& Future volumetric occupancy over $(t,x,y,z)$ \\

Temporal configuration
& $6$ future steps at $2$ Hz
& $4$ historical and $6$ future occupancy frames for occupancy prediction \\

Spatial discretization
& $64\,\mathrm{m}\times64\,\mathrm{m}$ ego-centered BEV; $1\,\mathrm{m}$ per cell
& $[-40,40]\times[-40,40]\times[-1,5.4]\,\mathrm{m}$; $0.4\,\mathrm{m}$ voxels \\

Semantic output
& $64\times64$ grid with $11$ categorical classes, including free space
& $200\times200\times16$ grid with $17$ semantic classes plus free space \\

World-state representation
& Latent shape $6\times8\times8\times8$ $(T\times H\times W\times C)$; $384$ occupancy tokens of dimension $8$
& $625$ history tokens jointly summarizing four historical 3D occupancy frames \\

Spatial route prediction
& Dedicated learnable route query decoded into $20$ spatial route points outside the joint flow; supervised directly
& Not used \\

Training procedure
& QA adaptation followed by joint Stage-I/II training with $\lambda_{\mathrm{II}}=0.5$
& QA adaptation followed by joint Stage-I/II training with $\lambda_{\mathrm{II}}=0.5$, followed by Stage-II output-decoder fine-tuning \\

Training objectives
& Latent flow matching, decoded occupancy reconstruction, trajectory consistency, and World--Action coupling
& Latent flow matching, ego-future prediction, and World--Action coupling \\

\bottomrule
\end{tabularx}
\end{table*}

Both benchmark configurations use a flow-time shift of $\rho=5$ and are trained with bfloat16 precision on eight A100 80GB GPUs.

\subsection{Stage-II Occupancy Modeling}
\label{sec:app_uncovered}

For both benchmarks, the scene-to-ego transformation produces the re-indexed occupancy $\bar O^E$ together with a validity mask $M^E$. The mask indicates whether each destination location is supported by the spatial extent of the Stage-I scene-centric prediction. Unsupported locations are filled with the free-space class to form a valid categorical input for $\mathcal E_{\mathrm{occ}}$.

We downsample $M^E$ to the occupancy-token resolution and obtain a token-wise coverage ratio $\mu\in[0,1]$. The pair $[\mu,1-\mu]$ is supplied as an additional condition to the World Expert and the ego-centric occupancy output decoder, i.e., the velocity readout $\Psi_o^E$. This condition allows the model to distinguish valid transformed content from uncovered source regions whose categorical values have been filled with free space.

\paragraph{Supervision of uncovered source regions.}
The coverage pair characterizes the \emph{Stage-II source}, not the validity of its supervision. In particular, $M^E$ is not a loss mask and does not turn uncovered cells into unlabeled targets. The ego-centric occupancy objective in Eq.~\eqref{eq:stage_flow_loss} uses all available valid native future-ego occupancy targets, including targets in regions uncovered by $\bar O^E$. Source coverage and target-label availability are separate notions: the same flow-matching objective applies to covered and uncovered source regions whenever corresponding supervision is valid. No distinct hole-filling loss or unobserved-as-free target is introduced.

At an uncovered source location, the free-space value is a categorical encoding placeholder, not a claim that the future world is free or that the target label is unknown. A valid future-ego target, if available, supplies the supervision independently of the source coverage. Thus Stage II can infer content absent from the Stage-I transformed source and also correct prediction errors and pose-induced spatial misalignment where the source is covered. Where the native annotation has no valid semantic target, the placeholder must not be interpreted as an additional free-space training label.

\paragraph{Evaluation protocol.}
Occupancy evaluation follows the standard benchmark protocol and does not use $M^E$ as an evaluation mask. The reported mIoU and IoU are computed over the standard native future-ego occupancy target rather than only over regions covered by the transformed Stage-I source.

\subsection{World--Action Coupling Objectives}

Both stages employ an auxiliary coupling objective between the generated occupancy and trajectory. In Stage I, the absolute trajectory is evaluated against the scene-centric occupancy. In Stage II, the accumulated ego-pose increments are evaluated against the completed ego-centric occupancy.

In both stages, the coupling objective penalizes trajectory regions that depart from the drivable surface or overlap with static and dynamic occupancy. The occupancy probabilities are detached when these terms are computed. Consequently, the coupling objectives update the Action Expert without directly encouraging the World Expert to remove obstacles from its predicted future.

\endgroup
\end{document}